\documentclass{article}

\usepackage{microtype}
\usepackage{graphicx}
\usepackage{subfigure}
\usepackage{booktabs} 

\usepackage{amsmath}
\usepackage{amsthm}
\usepackage{mathtools}
\usepackage{xargs}

\usepackage{hyperref}

\newtheorem{definition}{Definition}

\newcommandx*{\deltasquaremoment}[2][1=r_1,  2=r_2]{\ensuremath{\overline{ \delta_{#1} \, \delta_{#2} }}}
\newcommandx*{\deltasquaremomentone}[1][1=r]{\ensuremath{\bar{ \delta_{#1}^2}}}
\newcommandx*{\deltamoment}[1][1=r]{\ensuremath{\overline{ \delta_{#1}}}}
\newcommand{\ytrue}{\ensuremath{ y_{i,\text{true}} }}
\newcommandx*{\yregi}[2][1=i,2=r]{\ensuremath{\hat{y}_{#1,#2}}}
\newcommandx*{\yregfull}[1][1=r]{\ensuremath{\ytrue + \delta_{i,#1}}}
\newcommandx*{\Deltaij}[2][1=r_1, 2=r_2]{\ensuremath{\Delta_{#1,#2}}}

\usepackage[accepted]{icml2019}

\icmltitlerunning{Error correcting algorithms for noisy regressors}

\begin{document}

\twocolumn[
\icmltitle{Error Correcting Algorithms for Sparsely Correlated Regressors}


 
\icmlsetsymbol{equal}{*}

\begin{icmlauthorlist}
\icmlauthor{Andr\'es Corrada-Emmanuel}{dec}
\icmlauthor{Eddie Zahrebelski}{swoop}
\icmlauthor{Ed Pantridge}{swoop}
\end{icmlauthorlist}

\icmlaffiliation{dec}{Data Engines Corporation, Boston, USA}
\icmlaffiliation{swoop}{Swoop, Cambridge, USA}

\icmlcorrespondingauthor{Andr\'es Corrada-Emmanuel}{andres.corrada@dataengines.com}

\icmlkeywords{Ground Truth Inference, Error Detection, Error Correction, Machine Learning, ICML}

\vskip 0.3in
]



\printAffiliationsAndNotice{}  

\begin{abstract}
Autonomy and adaptation of machines requires that they be able to measure their own errors. 
We consider the advantages and limitations of such an approach when a machine has to measure 
the error in a regression task. How can a machine measure the error of regression 
sub-components when it does not have the ground truth for the correct predictions? A 
compressed sensing approach applied to the error signal of the regressors can recover their 
precision error without any ground truth.
It allows for some regressors to be \emph{strongly correlated} as long as not too many are so related.
Its solutions, however, are not unique - a property of ground truth inference solutions. Adding  $\ell_1$--minimization
as a condition can recover the correct solution
in settings where error correction is possible.
We briefly discuss the similarity of the mathematics of ground truth inference for regressors 
to that for classifiers.
\end{abstract}

\section{Introduction}
\label{introduction}
An autonomous, adaptive system, such as a self-driving car, needs to be robust to 
self-failures and changing environmental conditions. To do so, it must distinguish between self-errors and environmental changes. 
This chicken-and-egg problem is the concern of \emph{ground truth inference algorithms} - algorithms that measure a statistic of 
ground truth given the output of an ensemble of evaluators. They seek to answer the question - Am I malfunctioning or is 
the environment changing so much that my models are starting to break down?

Ground truth inference algorithms have had a spotty history in the machine learning community. The original idea came 
from \cite{Dawid79} and used the EM algorithm to solve a maximum-likelihood equation. This enjoyed a brief renaissance in the 
2000s due to advent of services like Amazon Mechanical Turk. Our main critique of all these approaches is that they are 
parametric - they assume the existence of a family of probability distributions for how the estimators are committing their errors. 
This has not worked well in theory or practice \cite{Zheng}. 

Here we will discuss the advantages and limitations of a non-parametric approach that uses compressed 
sensing to solve the  ground truth inference problem for noisy regressors \cite{CorradaSchultz2008}.
\emph{Ground truth} is defined in this context as the correct values for the predictions of the regressors. The existence of such 
ground truth is taken as a postulate of the approach. More formally,
\begin{definition}[Ground truth postulate for regressors]
All regressed values in a dataset can be written as,
\begin{equation}
\yregi = y_{i,\text{true}} + \delta_{i,r},
\end{equation}
where \ytrue \, does not depend on the regressor used.
\end{definition}
In many practical situations this is a very good approximation to reality. But it can be violated. For example, the regressors may 
have developed their estimates at different times. Meanwhile, $y(t)_{i,\text{true}}$ may have varied under them.

We can now define the ground truth inference problem for regressors as,
\begin{definition}[Ground truth inference problem for regressors]
Given the output of $R$ aligned regressors on a dataset of size D,
$$((\yregi[1][1], \yregi[2][1], \ldots, \yregi[D][1]),\ldots, (\yregi[1][R], \yregi[2][R], \ldots, \yregi[D][R])),$$
estimate the error moments for the regressors,
\begin{equation}
\deltasquaremoment{} \coloneqq  \frac{1}{D} \sum_{i=1}^D \delta_{i,r_1} \, \delta_{i,r_2},
\end{equation}
and
\begin{equation}
\deltamoment \coloneqq \frac{1}{D} \sum_{i=1}^D \delta_{i,r} ,
\end{equation}
without the true values, \{\ytrue\}.
\end{definition}
The separation of moment terms that are usually combined to define a covariance\footnote{To wit, covariance can
be expressed as $\sigma_{r_i,r_j} = \deltasquaremoment - \deltamoment[r_1]\,\deltamoment[r_2].$} between estimators is
 deliberate and relates to the math for the recovery as the reader will understand shortly. 

As stated, the ground truth inference problem for sparsely correlated regressors was solved in \cite{CorradaSchultz2008}
by using a compressed sensing approach to recover the $R(R+1)/2$ moments, \deltasquaremoment, for unbiased 
($\deltamoment \approx 0$)
regressors. Even the case of some of the regressors being strongly correlated is solvable. Sparsity of non-zero correlations is all that is required.
Here we point out that the failure to find a \emph{unique} solution for biased regressors still makes it possible to detect and correct 
biased regressors under the same sort of engineering logic that
allows bit flip error correction in computers. 

\section{Independent, unbiased regressors}

We can understand the advantages and limitations of doing ground truth inference for regressors by simplifying the problem to that of 
independent, un-biased regressors. The inference problem then becomes a straightforward linear algebra one that can be understood
 without  the complexity required when some unknown
number of them may be correlated.

Consider two regressors giving estimates,
\begin{align}
\yregi[i][r_1] = \yregfull[r_1]\\
\yregi[i][r_2] = \yregfull[r_2].
\end{align}
By the Ground Truth Postulate, these can be subtracted to obtain,
\begin{equation}
\label{eq:yregdiff}
\yregi[i][r_1] - \yregi[i][r_2] = \delta_{i,r_1} - \delta_{i,r_2}
\end{equation}
Note that the left-hand side involves \emph{observable} values that do not require any knowledge of \ytrue. The right hand side contains
 the error quantities that we seek to estimate. Squaring both sides and averaging over all the datapoints in the dataset we obtain our primary equation,
\begin{multline}
\label{eq:deltasquared}
\Deltaij^2 \coloneqq \frac{1}{D} \sum_{i=1}^D (\yregi[i][r_1] - \yregi[i][r_2])^2 \\
= \deltasquaremomentone[r_1] - 2\,\deltasquaremoment[r_1][r_2] + \deltasquaremomentone[r_2].
\end{multline}

Since we are assuming that the regressors are independent in their errors ($\deltasquaremoment[r_1][r_2] \approx 0$),
we can simplify \ref{eq:deltasquared} to,
\begin{equation}
 \deltasquaremomentone[r_1] + \deltasquaremomentone[r_2] = \Deltaij^2.
\end{equation}

This is obviously unsolvable for the unknown square moments, \deltasquaremomentone[r_1] and \deltasquaremomentone[r_2],
with a single pair of regressors. But for three regressors it is. It leads to the following linear algebra equation,
\begin{gather}
  \begin{bmatrix}
   1 & 1 & 0 \\
   1 & 0 & 1 \\
   0 & 1 & 1
   \end{bmatrix} \,
    \begin{bmatrix}
   \deltasquaremomentone[1] \\
   \deltasquaremomentone[2] \\
  \deltasquaremomentone[3]
   \end{bmatrix}
   =
   \begin{bmatrix}
   \Deltaij[1][2]^2 \\
   \Deltaij[1][3]^2 \\
   \Deltaij[2][3]^2
   \end{bmatrix}
\end{gather}

\begin{figure}
  \centering
  \begin{tabular}{@{}c@{}}
    \includegraphics[width=\linewidth]{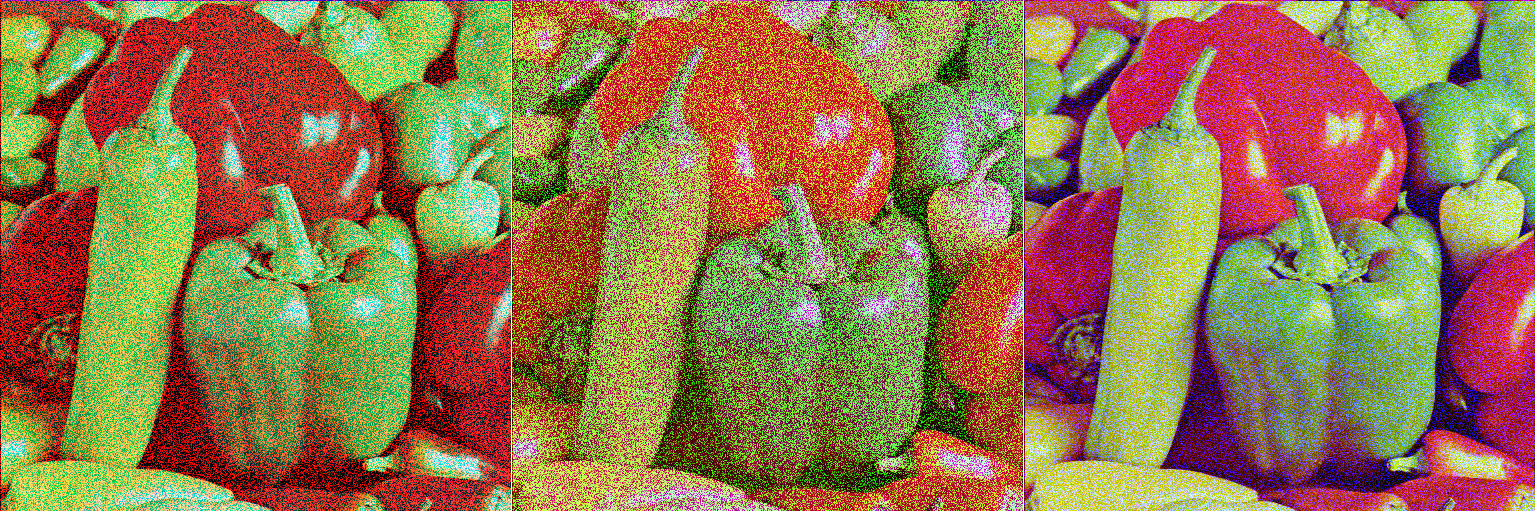} 
  \end{tabular}

  \begin{tabular}{@{}c@{}}
    \includegraphics[width=\linewidth]{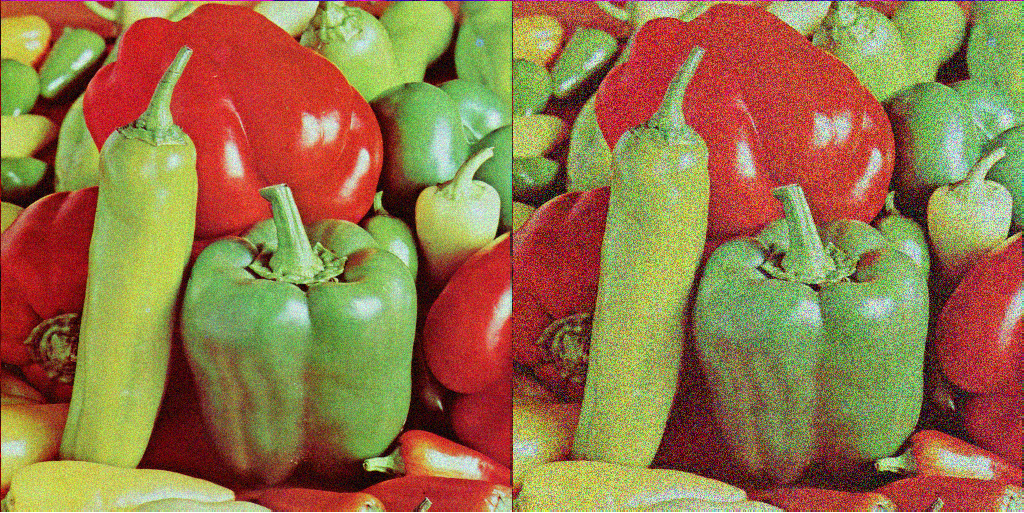} 
  \end{tabular}

  \caption{Least squares recovery of the error moments for three independent regressors. Three regressors
  with large noise in the red, green and blue channels respectively. Bottom left figure is the least squares
  solution used to weigh the channels. Bottom right figure is the simple averaging of the three regressors (Wisdom
  of the Crowd)}\label{fig:peppers}
\end{figure}

An application of this simple equation to a synthetic experiment with three noisy regressors is shown in 
Figure~\ref{fig:peppers}. Just like any least squares approach, and underlying topology for the relation between the
different data points is irrelevant. Hence, we can treat, for purposes of experimentation, each pixel value of a photo as a
ground truth value to be regressed by the synthetic noisy regressors. In this experiment we used uniform error. Similar
results are obtainable with other error distributions (e.g. Gaussian noise).

To highlight the multidimensional nature of equation 6, we randomized each of the color channels but made one channel
more noisy for each of the pictures. This simulates two regressors being mostly correct, but a third one perhaps
malfunctioning. 
Since even synthetic experiments with independent regressors will result in spurious non-zero cross-correlations, 
we solved the equation via least squares\footnote{The full compressive sensing solution being wholly unnecessary in this 
application where we know the regressors are practically independent.}. 


\section{Biased, independent regressors}

Our approach is not a typical one in statistics. Here we are concerned with the recovery of the multiple
error signals of the regressors, not the single underlying \{\ytrue\} signal. Wainwright (\cite{Wainwright})
discusses sparsity in the context of high-dimensional data to discuss the typical use of cormpressed
sensing - to recover the signal, not statistics of the many error signals of the regressors. Likewise, 
\emph{seemingly unrelated regressions} \cite{SUR} by Zellner is focused on recovering the model
parameters of a linear model for an ensemble of regressors.


Our approach fails for the case of \emph{biased} regressors. 
We can intuitively understand that because eq. \ref{eq:yregdiff} is invariant to a global bias, $\Delta$, for the regressors. 
We are not solving for the \emph{full average} error of the regressors but their \emph{average precision} error,
\begin{gather}
  \begin{bmatrix}
   \overline{\epsilon_1} \\
   \vdots \\
   \overline{\epsilon_r}
   \end{bmatrix}
   =
    \begin{bmatrix}
   \overline{\delta_1} \\
   \vdots \\
   \overline{\delta_r}
   \end{bmatrix}
   +
   \begin{bmatrix}
   \Delta \\
   \vdots \\
   \Delta
   \end{bmatrix}
\end{gather}
We can only determine the error of the regressors modulus some unknown global bias.
This, by itself, would not be an unsurmountable problem since global shifts are easy to fix. From an engineering 
perspective, accuracy is cheap while precision is expensive\footnote{Examples are (a) the zeroing screw in a 
precision weight scale, (b) the number of samples needed to measure a classifier's accuracy when it is of unknown 
accuracy versus when we know it is either, say,  1\% or 99\% accurate. The former situation is more accurate on 
average but less precise. The latter one, precise but inaccurate.}. The more problematic issue is that it would not 
be able to determine correctly who is biased if they are biased \emph{relative} to each other.

Let us demonstrate that by using eq \ref{eq:yregdiff} to estimate the average bias, \deltamoment[r], for the regressors.
 Averaging over both sides, we obtain for three independent regressors, the following equation\footnote{The observable
statistic, $\Delta_{r_1,r_2}$, is equal to $1/D \sum (\yregi[i][r_1] - \yregi[i][r_2]).$},
\begin{gather}
  \begin{bmatrix}
   1 & -1 & 0 \\
   1 & 0 & -1 \\
   0 & 1 & -1
   \end{bmatrix} \,
    \begin{bmatrix}
   \deltamoment[1] \\
   \deltamoment[2] \\
   \deltamoment[3]
   \end{bmatrix}
   =
   \begin{bmatrix}
   \Deltaij[1][2] \\
   \Deltaij[1][3] \\
   \Deltaij[2][3]
   \end{bmatrix}
\end{gather}
The rank of this matrix is two. This means that the matrix has a one-dimensional null space. In this particular case, 
the subspace is spanned by a constant bias shift as noted previously. Nonetheless, let us consider the specific case of 
three regressors where two of them have an equal constant bias,
\begin{gather}
\begin{bmatrix}
\deltamoment[1] \\
 \deltamoment[2] \\
 \deltamoment[3]
\end{bmatrix}
=
\begin{bmatrix}
0 \\
 \Delta \\
 \Delta
\end{bmatrix}.
\end{gather}
This would result in the $\Deltaij$ vector,
\begin{gather}
\begin{bmatrix}
\Deltaij[1][2] \\
\Deltaij[1][3] \\
\Deltaij[2][3]
\end{bmatrix}
=
\begin{bmatrix}
-\Delta \\
-\Delta \\
0
\end{bmatrix}.
\end{gather}
The general solution to Eq.\ 10 would then be,
\begin{gather}
\begin{bmatrix}
\deltamoment[1] \\
 \deltamoment[2] \\
 \deltamoment[3]
\end{bmatrix}
=
\begin{bmatrix}
-\Delta \\
0 \\
0
\end{bmatrix}
+
c
\begin{bmatrix}
\Delta \\
\Delta \\
\Delta
\end{bmatrix}.
\end{gather}
This seems to be a failure for any ground truth inference for noisy regressors. Lurking underneath this math is the core idea of 
compressed sensing: pick the value of $c$ for the solutions to eq.\ 14 that minimizes the $\ell_1$ norm of the recovered 
vector. When such a point of view is taken, non-unique solutions to ground truth inference problems can be re-interpreted as 
error detecting and correcting algorithms. We explain.

\section{Error detection and correction}

Suppose, instead, that only one of the three regressors was biased,
\begin{gather}
\begin{bmatrix}
\deltamoment[1] \\
 \deltamoment[2] \\
 \deltamoment[3]
\end{bmatrix}
=
\begin{bmatrix}
\Delta \\
 0 \\
 0
\end{bmatrix}.
\end{gather}
This would give the general solution,
\begin{gather}
\begin{bmatrix}
\deltamoment[1] \\
 \deltamoment[2] \\
 \deltamoment[3]
\end{bmatrix}
=
\begin{bmatrix}
\Delta \\
0 \\
0
\end{bmatrix}
+
c
\begin{bmatrix}
\Delta \\
\Delta \\
\Delta
\end{bmatrix},
\end{gather}
with $c$ an arbitrary, constant scalar.
If we assume that errors are sparse, then an $\ell_1$-minimization approach would lead us to select the solution,
\begin{gather}
\begin{bmatrix}
\deltamoment[1] \\
 \deltamoment[2] \\
 \deltamoment[3]
\end{bmatrix}
=
\begin{bmatrix}
\Delta \\
0 \\
0
\end{bmatrix}.
\end{gather}
The algorithm would be able to detect and correct the bias of a single regressor. If we wanted more reassurance that we were 
picking the correct solution then we could use 5 regressors. When the last two have constant bias, the general solution is,
\begin{gather}
\begin{bmatrix}
\deltamoment[1] \\
\deltamoment[2] \\
\deltamoment[3] \\
\deltamoment[4] \\
\deltamoment[5] \\
\end{bmatrix}
=
\begin{bmatrix}
-\Delta \\
-\Delta \\
-\Delta \\
0 \\
0 \\
\end{bmatrix}
+
c
\begin{bmatrix}
\Delta \\
\Delta \\
\Delta \\
\Delta \\
\Delta
\end{bmatrix}.
\end{gather}
With the corresponding $\ell_1$-minimization solution of,
\begin{gather}
\begin{bmatrix}
\deltamoment[1] \\
\deltamoment[2] \\
\deltamoment[3] \\
\deltamoment[4] \\
\deltamoment[5] \\
\end{bmatrix}
=
\begin{bmatrix}
0 \\
0 \\
0 \\
\Delta \\
\Delta \\
\end{bmatrix}.
\end{gather}
This is the same engineering logic that makes practical the use of
error correcting codes when transmitting a signal over a noisy channel. Our contribution is to point out that 
the same logic \emph{also} applies to \emph{estimating errors} by regressors trying to recover the true signal.

\section{Conclusions}

A compressed sensing algorithm for recovering the average error moments of an ensemble of noisy regressors exists. 
Like other ground truth inference algorithms, it leads to non-unique solutions. However, in many well-engineered systems, 
errors are sparse and mostly uncorrelated when the machine is operating normally. Algorithms such as this one can then 
detect the beginning of malfunctioning sensors and algorithms.

\begin{figure}
  \centering
    \includegraphics[width=\linewidth]{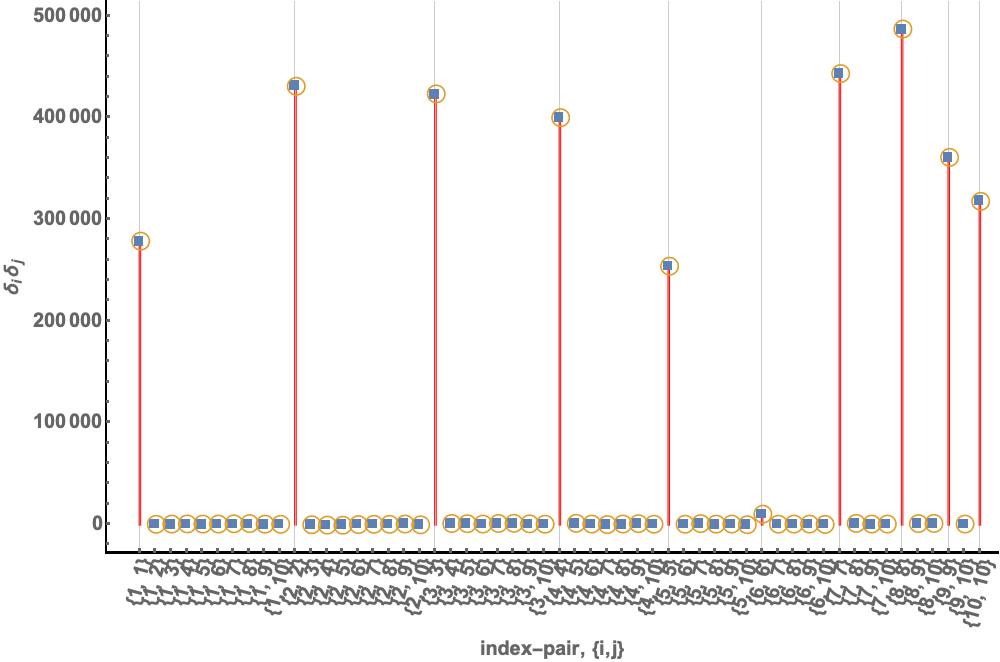} 
    \vskip -0.2in
  \caption{Recovered square error moments (circles), \deltasquaremoment, for the true error moments (squares) of 10 
  synthetic regressors on the pixels of a 1024x1024 image. Recovering algorithm does not know which vector components
  correspond to the strong diagonal signal, the (i,i) error moments.}\label{fig:recovered-moments}
  \vskip -0.2in
\end{figure}

We can concretize the possible applications of this technique by considering a machine such as a self-driving car. 
Optical cameras and range finders are necessary sub-components. How can the car detect a malfunctioning sensor? 
There are many ways this already can be done (no power from the sensor, etc.). This technique adds another layer of 
protection by potentially detecting anomalies earlier. In addition, it allows the creation of \emph{supervision} arrangements 
such as having one expensive, precise sensor coupled with many cheap, imprecise ones. As the recovered error moment 
matrix in Figure~\ref{fig:recovered-moments} shows, many noisy sensors can be used to benchmark a more precise 
one (the (sixth regressor \{6,6\} moment in this particular case).
As \cite{CorradaSchultz2008} demonstrate, it can also be used on the final output of algorithms. In the case of a self-driving car, 
a depth map is needed of the surrounding environment - the output of algorithms processing the sensor input data. 
Here again, one can envision supervisory arrangements where quick, imprecise estimators can be used to monitor a more 
expensive, precise one.

There are advantages and limitations to the approach proposed here.
Because there is no maximum likelihood equation to solve, the method is widely applicable. The price for this flexibility
is that no generalization can be made. There is no theory or model to explain the observed errors - they are just
estimated robustly for each specific dataset. Additionally, the math is easily understood. The advantages or limitations of a 
proposed application to an autonomous, adaptive system can be ascertained readily. 
The theoretical guarantees of compressed sensing algorithms are a testament to this \cite{Foucart}.
Finally, the compressed sensing approach to regressors can handle \emph{strongly}, but \emph{sparsely}, 
correlated estimators.

We finish by pointing out that non-parametric methods also exist for classification tasks. This is demonstrated for independent, binary 
classifiers (with working code) in \cite{Corrada2018}. The only difference is that the linear algebra of the regressor problem 
becomes polynomial algebra. Nonetheless, there we find similar ambiguities due to non-unique solutions to the ground truth 
inference problem of determining average classifier accuracy without the correct labels. For example, the polynomial for unknown
 prevalence (the environmental variable) of one of the labels is quadratic, leading to two solutions. Correspondingly, the accuracies 
 of the classifiers (the internal variables) are either $x$ or $1-x$. So a single classifier could be, say, 90\%  or 10\% accurate. 
 The ambiguity is removed by having enough classifiers - the preferred solution is where one of them is going below 50\%, not the rest doing so.

\bibliography{data-engines-amtl2019}

\begin{thebibliography}{7}
\providecommand{\natexlab}[1]{#1}
\providecommand{\url}[1]{\texttt{#1}}
\expandafter\ifx\csname urlstyle\endcsname\relax
  \providecommand{\doi}[1]{doi: #1}\else
  \providecommand{\doi}{doi: \begingroup \urlstyle{rm}\Url}\fi

\bibitem[Corrada-Emmanuel(2018)]{Corrada2018}
Corrada-Emmanuel, A.
\newblock Ground truth inference of binary classifier accuracies - the
  independent classifiers case.
\newblock
  \url{https://github.com/andrescorrada/ground-truth-problems-in-business/blob/master/classification/IndependentBinaryClassifiers.pdf},
  2018.

\bibitem[Corrada-Emmanuel \& Schultz(2008)Corrada-Emmanuel and
  Schultz]{CorradaSchultz2008}
Corrada-Emmanuel, A. and Schultz, H.
\newblock Geometric precision errors in low-level computer vision tasks.
\newblock In \emph{Proceedings of the 25th International Conference on Machine
  Learning (ICML 2008)}, pp.\  168--175, Helsinki, Finland, 2008.

\bibitem[Dawid et~al.(1979)Dawid, Skene, Dawidt, and Skene]{Dawid79}
Dawid, P., Skene, A.~M., Dawidt, A.~P., and Skene, A.~M.
\newblock Maximum likelihood estimation of observer error-rates using the em
  algorithm.
\newblock \emph{Applied Statistics}, pp.\  20--28, 1979.

\bibitem[Foucart \& Rauhaut(2013)Foucart and Rauhaut]{Foucart}
Foucart, S. and Rauhaut, H.
\newblock \emph{A Mathematical Introduction to Compressive Sensing}.
\newblock Birkh{\"a}user, New York, 2013.

\bibitem[Wainwright(2019)]{Wainwright}
Wainwright, M.~J.
\newblock \emph{High-Dimensional Statistics: A Non-Asymptotic Viewpoint}.
\newblock Cambridge University Press, New York, 2019.

\bibitem[{Wikipedia contributors}(2019)]{SUR}
{Wikipedia contributors}.
\newblock Seemingly unrelated regressors --- {W}ikipedia{,} the free
  encyclopedia, 2019.
\newblock URL
  \url{https://en.wikipedia.org/wiki/Seemingly_unrelated_regressions}.
\newblock [Online; accessed 3-July-2019].

\bibitem[Zheng et~al.(2017)Zheng, Li, Li, Shan, and Cheng]{Zheng}
Zheng, Y., Li, G., Li, Y., Shan, C., and Cheng, R.
\newblock Truth inference in crowdsourcing: Is the problem solved?
\newblock In \emph{Proceedings of the VLDB Endowment}, volume 10, no. 5, 2017.

\end{thebibliography}
\bibliographystyle{icml2019}
\end{document}